%% file: main.tex
\RequirePackage{fix-cm}
\documentclass{svjour3}                     

\usepackage{amssymb}
\usepackage{amsmath,graphicx}
\usepackage{amssymb}
\usepackage{algorithm}
\usepackage{algpseudocode}
\usepackage{multirow}
\usepackage{array}
\usepackage{url}
\usepackage[super]{natbib}
\usepackage{color}
\usepackage{comment}
\smartqed  
\usepackage{graphicx}
\usepackage{lscape} 
\usepackage{import}
\usepackage{caption}
\usepackage{subcaption}
\usepackage{xcolor}
\usepackage{amsmath}
\usepackage[backref=page]{hyperref}
\usepackage[left]{lineno} 
 \journalname{IPCAI}
\begin{document}

\title{Simulation-to-Real domain adaptation with teacher-student learning for endoscopic instrument segmentation   \thanks{Funded by the German Federal Ministry of Education and Research (BMBF) under the project COMPASS (grant no. - 16\,SV\,8019).}
}

\titlerunning{Simulation-to-Real}        

\author{Manish Sahu \and
        Anirban Mukhopadhyay \and 
        Stefan Zachow
        }

\authorrunning{Sahu et al.} 

\institute{{M. Sahu \and S. Zachow} \at
              Zuse Institute Berlin (ZIB), Germany \\
              \email{\{sahu, zachow\}@zib.de}        
           \and
           A. Mukhopadhyay \at
              Department of Computer Science, TU Darmstadt, Germany \\
              \email{anirban.mukhopadhyay@gris.tu-darmstadt.de}
}

\date{Received: date / Accepted: date}

\maketitle


\begin{abstract}

\textbf{\newline \\* Purpose}:
Segmentation of surgical instruments in endoscopic videos is essential for automated surgical scene understanding and process modeling. 
However, relying on fully supervised deep learning for this task is challenging because manual annotation occupies valuable time of the clinical experts.
\smallskip 
\textbf{\\* Methods}: 
We introduce a teacher-student learning approach that learns jointly from annotated simulation data and unlabeled real data to tackle the erroneous learning problem of the current consistency-based unsupervised domain adaptation framework.
\smallskip 
\textbf{\\* Results}: 
Empirical results on three datasets highlight the effectiveness of the proposed framework over current approaches for the endoscopic instrument segmentation task.
Additionally, we provide analysis of major factors affecting the performance on all datasets to highlight the strengths and failure modes of our approach.
\smallskip 
\textbf{\\* Conclusion}: 
We show that our proposed approach can successfully exploit the unlabeled real endoscopic video frames and improve generalization performance over pure simulation-based training and the previous state-of-the-art.
This takes us one step closer to effective segmentation of surgical tools in the annotation scarce setting.



\end{abstract}

\section{Introduction} \label{intro}
\input{Sections/Introduction}

\section{Related Work} \label{sec:rel_work}
\input{Sections/RelatedWork}

\section{Method} \label{sec:method}
\input{Sections/Method}

\section{Experimental Setup} \label{sec:experiments}
\input{Sections/Experiments}


\section{Results and Discussion} \label{sec:results}
\input{Sections/Results}


\section{Conclusion} \label{sec:disc}
\input{Sections/Discussion}


%
%
\input{Sections/COI}

\bibliographystyle{spbasic}      
\bibliography{Bibliography}


\end{document}

%% file: Sections/Introduction.tex
A faithful segmentation of surgical instruments in endoscopic videos is a crucial component of surgical scene understanding and realization of automation in computer- or robot-assisted intervention systems.\footnote{EndoVis Sub-challenges - 2015, 2017, 2018, 2019 [https://endovis.grand-challenge.org]}
A majority of recent approaches address the problem of surgical instrument segmentation by training deep neural networks (DNNs) in a fully-supervised scheme.
However, the applicability of such supervised approaches is restricted by the availability of a sufficiently large amount of real videos with clean annotations.
The annotation process (especially pixel-wise) can be prohibitively expensive (see Figure~\ref{fig:annotation_time}) because it takes valuable time of medical experts.

\begingroup
\begin{figure}[b]
    \centering
    \includegraphics[width=\textwidth]{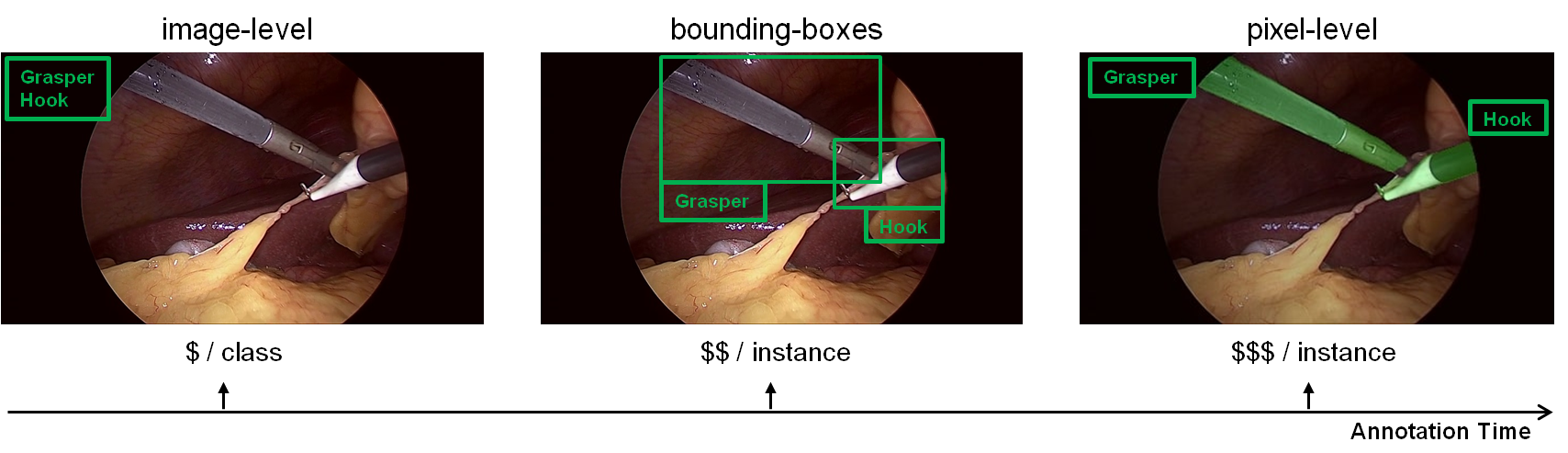}
	\caption{Fully supervised deep learning is unrealistic for instrument segmentation due to a significantly high annotation effort.}
    \label{fig:annotation_time}
\end{figure}
\endgroup

An alternative direction to mitigate the dependency on annotated video sequences is to utilize synthetic data for the training of DNNs.
Recent advances in graphics and simulation infrastructures have paved the way to automatically create a large number of photo-realistic simulated images with accurate pixel-level labels.\cite{hoffman2016fcns,pfeiffer2019generating}
However, the DNNs trained purely on simulated images do not generalize well on real endoscopic videos due to domain shift issue.\cite{torralba2011unbiased,vercauteren2019cai4cai}
We hypothesize that a DNN's bias towards recognizing textures rather than shapes~\cite{baker2018deep} results in a significant drop of performance when the DNNs are trained on simulation (rendered) data and applied to real environments.
This is mainly because the heterogeneity of information within a real surgical scene is heavily influenced by factors such as lighting conditions, motion blur, blood, smoke, specular reflection, noise etc. However, simulation data only mimic shapes of instrument and patient-specific organs. \cite{engelhardt2019cross}
To alleviate this performance generalization issue, domain adaptation approaches\cite{wang2018surveyda} are utilized to address domain shift/mismatch.

When the annotations for the target domain are not available for learning, domain adaptation is applied in an unsupervised manner,\cite{wilson2020surveyuda} where enabling the DNN to learn domain-invariant features is the primary goal of a learning algorithm.
For instance, Pfeiffer et al.~\cite{pfeiffer2019generating} utilized an \emph{image-to-image} translation approach, where the simulated images are translated into realistic looking ones by mapping image styles (texture, lighting) of the real data using a Cycle-GAN.
In contrast, we argued for a shape-focused joint learning from simulated and real data in an end-to-end fashion and introduced a consistency-learning based approach \emph{Endo-Sim2Real}\cite{sahu2020endo} to align the DNNs on both domains.
We showed that similar performance can be obtained by employing a non-adversarial approach while improving the computational efficiency (with respect to training time).
However, similar to perturbation-based consistency learning approaches,~\cite{laine2016temporal,french2018self} \emph{Endo-Sim2Real} suffers from so-called confirmation bias.\cite{tarvainen2017mean} This is caused by noise accumulation or erroneous learning during the training stages, which may result in a degenerate solution.\cite{chapelle2009semi}

In this work, we formalise the consistency-based unsupervised domain adaptation framework to identify the confirmation bias problem of \emph{Endo-Sim2Real} and propose a teacher-student learning paradigm to address this problem.
Our proposed approach tackles the erroneous learning by improving the pseudo-label generation procedure for the unlabeled data and facilitate stable training of DNNs while maintaining computational efficiency.
Through quantitative and qualitative analysis, we show that our proposed approach outperforms the previous \emph{Endo-Sim2Real} approach across three data sets. Moreover, the proposed approach leads to a stable training without loosing computational efficiency.

\noindent Our contribution in this article is two-fold:
\begin{enumerate}
    \item We introduce the teacher-student learning paradigm to the task of surgical instrument segmentation in endoscopic videos.
    \item We provide a thorough quantitative and qualitative analysis to show the failure modes of our unsupervised domain adaptation approach. Our analysis leads to a better understanding of the strengths and challenges of consistency-based unsupervised learning with simulation-based supervision.
\end{enumerate}

%% file: Sections/RelatedWork.tex
Research on instrument segmentation for endoscopic procedures is dominated by supervision-based approaches ranging from full supervision\cite{bodenstedt2018comparative}, semi/self supervision\cite{ross2018exploiting}, and weak supervision~\cite{fuentes2019easylabels} up to multi-task~\cite{laina2017concurrent} and multi-modal learning\cite{jin2019incorporating}.
Some recent works also explored unsupervised approaches,\cite{colleoni2020synthetic,liu2020unsupervised} however, for the sake of brevity, we will only focus on  approaches that employ learning from simulation data for unsupervised domain adaptation.

Within the context of domain adaptation in surgical domains, Mahmood et. al.~\cite{mahmood2018unsupervised} proposed an adversarial-based  transformer network to translate a real image to a synthetic image such that a depth estimation model trained on synthetic images can be applied to the real image.
On the other hand, Rau et. al.~\cite{rau2019implicit} proposed a conditional Generative Adversarial Network (GAN) based approach to estimate depth directly from real images.
Other works have argued for translating synthetic images to photo-realistic images by using domain mapping via style transfer\cite{luengo2018surreal,marzullo2020towards}, for instance by using Cycle-GAN based \emph{unpaired image-to-image} translation\cite{engelhardt2018improving,oda2019realistic} and utilize annotations from synthetic environment for deep learning tasks.

Pfeiffer et.~al.~\cite{pfeiffer2019generating} proposed an unpaired \textit{image-to-image} translation approach \textit{I2I}  that focuses on reducing the distribution difference between the source and the target domain by employing a Cycle-GAN based style transfer. Afterwards, a DNN is trained on the the translated images and its corresponding labels.
On the other hand, \emph{Endo-Sim2Real}~\cite{sahu2020endo} utilizes similarity-based joint learning from both simulation and real data under the assumption that the shape of an instrument remains consistent across domains as well as under semantic preserving perturbations (like adding pixel-level noise or transformations).

This work is in line with \emph{Endo-Sim2Real} and focuses on end-to-end learning for unsupervised domain adaptation (UDA).
However, we formalise the consistency-based UDA to identify the confirmation bias problem and unstable training of \emph{Endo-Sim2Real} approach and address it by employing a teacher-student paradigm.
This facilitates stable training of the DNN and enhances its performance generalization capability.

%% file: Sections/Method.tex

\begingroup
\begin{figure}[t]
    \centering
    \includegraphics[width=\textwidth]{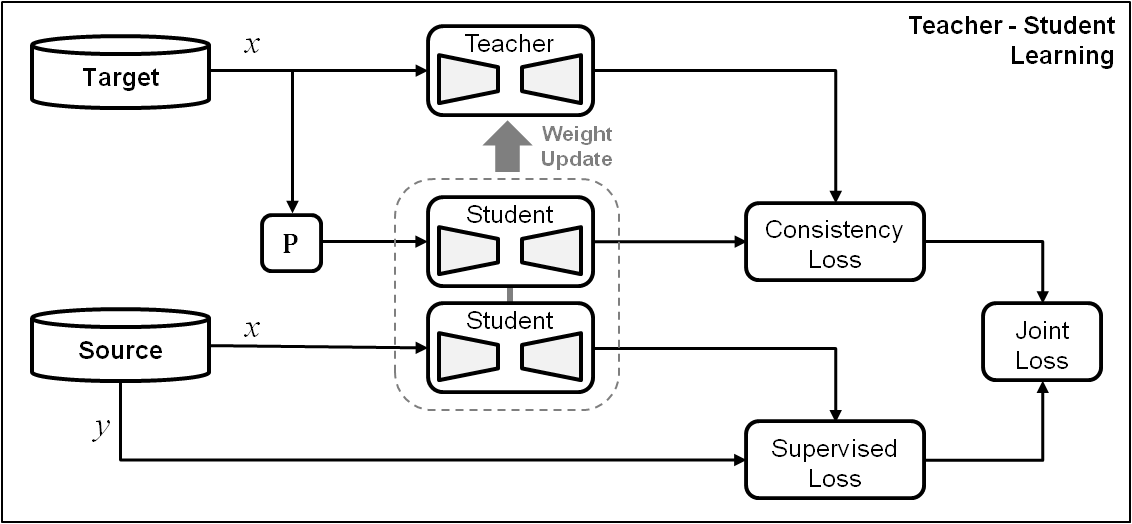}
	\caption{Proposed teacher-student learning approach comprising supervised learning from simulated data as well as consistency learning from real unlabeled data.}
    \label{method:pipeline}
\end{figure}
\endgroup

Our proposed teacher-student domain-adaptation approach (see Figure~\ref{method:pipeline}) aims to bridge the domain gap between source (simulated) and target (real) data by aligning a DNN model to both domains.
Given:
\begin{itemize}
    \item[$\bullet$] a  source  domain $D_s = (X_s,Y_s)$ associated with a feature space $X_s$ and a label space $Y_s$ and containing $n_s$ labeled samples $\{ (x_i^s, y_i^s) \}_{i=1}^{n_s}$ where, $x_i \in X_s$ and $y_i \in Y_s$ denote the $i$-th pair of image and label data respectively
    \item[$\bullet$] a target domain $D_t = (X_t)$  associated with a feature space $|X_t|$ and  a label space $|Y_t|$ and containing $n_t$ unlabeled samples $\{ x_i^t \}_{i=1}^{n_t}$ where, $x_i \in X_t$ denote the $i$-th image of the unlabeled data
\end{itemize}

\noindent the goal of unsupervised domain adaptation is to learn a DNN model that generalizes on the target domain $D_t$. 
It is important to note that although the simulation and real endoscopic scene may appear similar, the label space between source- and target-domains generally differ (i.e. $Y_s \neq Y_t$), representing for example  different organs or different instrument types. 
Since we are focusing on binary instrument segmentation, the label categories are twofold (i.e. $Y_s = Y_t = \{ ``instrument", ``background" \}$).
For the sake of simplicity, we refer to the source domain $D_s$ as labeled simulated domain $D_L^{Sim}$ and to the target domain $D_t$ as unlabeled real domain $D_{UL}^{Real}$.

Our proposed (and previous) approach learns by jointly minimizing the supervised loss $L_{sl}$ for the labeled simulated data-pair as well as the consistency loss $L_{cl}$ for the unlabeled real data.
A core component of the joint learning approach is unsupervised consistency learning, where a supervisory signal is generated by enforcing the DNN $f_{\theta}$ (parameterized with network weights $\theta$) to produce a consistent output for an unlabeled input $x$ and its perturbed form $\mathcal{P}(x)$.

\begin{equation}\label{Consistency Equation}
\min_{\theta}  \ \ \mathcal{L}_{sl} + \mathcal{L}_{cl} \ \  \big\{ \underbrace{f_{\theta} \big( x \big)}_{\tilde{\textbf{y}}}, f_{\theta} \big( \mathcal{P}(x) \big) \big\}
\end{equation}

Here in Equation~\ref{Consistency Equation}, the DNN prediction $\tilde{\textbf{y}}$ for unperturbed data $x$ acts as a \emph{pseudo-label} for perturbed data $\mathcal{P}(x)$ to guide the learning process.
Therefore, the \emph{Endo-Sim2Real} scenario can be interpreted as a \emph{student-as-teacher} approach where the DNN acts as both a teacher that produces pseudo-labels and a student that learns from these labels.
Since the DNN predictions may be incorrect or noisy during training,\cite{laine2016temporal} this \emph{student-as-teacher} approach leads to so-called the confirmation bias,\cite{tarvainen2017mean} which 
reinforces the student to overfit to the incorrect pseudo-labels generated by the teacher and prevents learning new information. This issue is especially prominent during early stages of the training, when the DNN still lacks the correct interpretation of the labels.
If the unsupervised consistency loss ($L_{cl}$) outweighs the supervised loss ($L_{sl}$), the learning process is not effective and leads to a sub-optimal performance. Therefore, the consistency loss is typically employed with a temporal weighting function $w(t)$ such that the DNN learns prominently from the supervised loss during the initial stages of the learning and gradually shifts towards unsupervised consistency learning in the later stages. 

Although the temporal ramp-up weighting function in \emph{Endo-Sim2Real} helps to reduce the effect of the confirmation bias during joint learning, the DNN still learns directly from the incorrect pseudo-labels generated by the teacher.

\input{Algorithms/pseudo_code}

In this work, we address this major drawback of the \emph{Endo-Sim2Real} approach by improving the pseudo-label generation procedure of the unlabeled consistency learning. To this end, the teacher network is de-coupled from the student network and redefined ($f_{\theta} \longrightarrow f_{\theta^{'}}$) to generate reliable targets to enable the student to gradually learn meaningful information about the instrument shape. In order to avoid separate training of the teacher model, the same architecture is used for the teacher and its parameters are updated as a temporal average~\cite{tarvainen2017mean} of the student network's weights.

\begin{equation*}\label{Loss Function}
\min_{\theta} \Bigg\{ \underbrace{\sum_{(x_i,y_i) \in D_L^{Sim}}  \mathcal{L}_{sl} \Big( f_{\theta} \big( x_i \big),y_i \Big) }_{supervised\: simulation}  + \ \  w(t) \ast \underbrace{ \sum_{x_i \in D_{UL}^{Real}}  \mathcal{L}_{cl} \Big( f_{\theta^{'}} \big( x_i \big), f_{\theta} \big( \mathcal{P}(x_i) \big) \Big) }_{unsupervised\: real} \Bigg\} 
\end{equation*}

\begin{equation}\label{EMA Equation}
\theta_{t}^{'} =  ( \alpha \cdot \theta_{t-1}^{'} + (1 - \alpha) \cdot \theta_{t} )
\end{equation}

At each training step $t$, the student $f_{\theta}$ is updated using \textit{gradient-descent} while the teacher $f_{\theta^{'}}$ is updated using student network weights, where the smoothing factor $\alpha$ controls the update rate of the teacher.
A pseudo code of our proposed teacher-student learning approach is provided in Algorithm~\ref{alg:domain_adaptation}.


%% file: Algorithms/pseudo_code.tex
\begingroup
\begin{algorithm}[htbp]
\caption{Teacher-student Algorithm for Domain Adaptation}\label{alg:domain_adaptation}
\begin{algorithmic}[1] 
    \renewcommand{\algorithmicrequire}{\textbf{Model:}}
    \Require $f_{\theta}$ \Comment{student with trainable parameters $\theta$.} 
    \Require $f_{\theta^{'}}$ \Comment{teacher with trainable parameters $\theta^{'}$} 
    \renewcommand{\algorithmicensure}{\textbf{Data:}}
    \Ensure $D_L^{Sim}(x, y)$ \Comment{labeled simulated samples}
    \Ensure $D_{UL}^{Real}(x)$ \Comment{unlabeled real samples}
    \renewcommand{\algorithmicrequire}{\textbf{Require:}}
    \Require $\alpha$ \Comment{update rate of teacher}
    \Require $w(t)$ \Comment{temporal weight of consistency loss}
    \renewcommand{\algorithmicensure}{\textbf{Ensure:}}
    \Ensure $\theta^{'}$ $\leftarrow$ $\theta$ \Comment{Initialize weights}
    \renewcommand{\algorithmicprocedure}{\textbf{loop}}
    \Procedure{Joint Learning}{$D_L^{Sim}, D_{UL}^{Real}$} \Comment{($D_s \rightarrow D_L^{Sim}$, $D_t \rightarrow D_{UL}^{Real}$)}
    \State \textbf{Supervised Loss}
    \State $\{ (x_i, y_i) \}_{i=1}^B \sim D_L^{Sim} \left( x, y \right)$ \Comment{Sample mini-batch}
    \State $\{ ( x_i^a, y_i^a) \}_{i=1}^B = \{ \mathcal{A} \left( x_i, y_i \right) \}_{i=1}^B$ \Comment{Augment batch}
    \State $L_{sl}$ $=$ $\{  f_{\theta}(x_i^a), y_i^a \}_{i=1}^B$ \Comment{Supervised loss}
    %
    \State \textbf{Consistency Loss}
    \State $\{x_i \}_{i=1}^B \sim D_{UL}^{Real}(x)$ \Comment{Sample mini-batch}
    \State $\{ x_i^p \}_{i=1}^B = \{ \mathcal{P} \left( x_i \right) \}_{i=1}^B$ \Comment{Perturb batch}
    \State $\{ \tilde{y}_i \}_{i=1}^B = \{ f_{\theta^{'}} \left( x_i^p \right) \}_{i=1}^B$ \Comment{Pseudo Segmentation}
    \State $L_{cl}$ $=$ $ \{  f_{\theta} \left( x_i \right), \tilde{y}_i \}_{i=1}^B$ \Comment{Unsupervised loss}
    \State \textbf{Joint Loss}
    \State $L = L_{sl} + w(t) \cdot L_{cl}$ \Comment{Joint loss}
    \State $g_{\theta} = {\nabla}_{\theta}L$ \Comment{Compute gradients}
    \State $\theta \gets  Update(\theta, g_{\theta})$ \Comment{Update student (gradient descent)}
    \State $\theta^{'} \gets  ( \alpha \cdot \theta^{'} + (1 - \alpha) \cdot \theta ) $ \Comment{Update teacher}
    \EndProcedure
    \newline \textbf{return} $\theta^{'}$\Comment{Learned model}
\end{algorithmic}
\end{algorithm}
\endgroup

%% file: Sections/Experiments.tex

\subsection{Data}
\label{dataset} 

\paragraph{\textbf{Simulation}~\cite{pfeiffer2019generating}}
data contain $20K$ rendered images acquired via 3-D laparoscopic simulations from the CT scans of 10 patients.
The images describe a rendered view of a laparoscopic scene with each tissue having a distinct texture and a presence of two conventional surgical instruments (grasper and hook) under a random placement of the camera (coupled with a light source).

\paragraph{\textbf{Cholec}~\cite{sahu2020endo}}
data contain around $7K$ endoscopic video frames acquired from 15 videos of the Cholec80 dataset.\cite{twinanda2016endonet}
The images describe the laparoscopic cholecystectomy scene with seven conventional surgical instruments (grasper, hook, scissors, clipper, bipolar, irrigator and specimen bag).
The data provide segmentations for each instrument type, however, the specimen bag is considered as a counterexample that is treated as background during evaluation, following the definition of an instrument in RobustMIS challenge.\cite{ross2020robust}

\begingroup
\begin{table}[htbp] 
    \subimport{./Tables/}{Datasets.tex}%
\end{table}
\endgroup

\paragraph{\textbf{EndoVis}~\cite{endovis2015}}
data consist of 300 images from six different in-vivo 2D recordings of complete laparoscopic colorectal surgeries. 
The data provide binary segmentations of instruments for validation where images describe an endoscopic scene containing seven conventional instruments (including hook, traumatic  grasper,  ligasure,  stapler,  scissors and scalpel).\cite{bodenstedt2018comparative}

\paragraph{\textbf{RobustMIS}~\cite{ross2020robust}}
data consist of around $10K$ images acquired from 30 surgical procedures of three different types of colorectal surgery (10 rectal resection procedures, 10 proctocolectomy procedures and 10 procedures of sigmoid resection procedures).
An instrument is defined as an elongated rigid object that is manipulated directly from outside the patient. Therefore, 
grasper, scalpel, clip applicator, hooks, stapling device, suction and even trocar is considered as an instrument while non-rigid tubes, bandages, compresses, needles, coagulation sponges, metal clips etc. are considered as counterexamples as they are indirectly manipulated from outside.\cite{ross2020robust}
The data provide instance level segmentations for validation, which are performed in three different stages with an increasing domain gap between the training- and the test-data. 
Stage 1 contains video frames from 16 cases of the training data, stage 2 has video frames of two proctocolectomy and rectal surgeries each, and stage 3 has video frames from 10 sigmoid resection surgeries.


\paragraph{It} is important to note that the domain gap increases not only in the three stages of testing in Robust-MIS dataset, but also from \textit{Simulation} towards \textit{Real} datasets (EndoVis $<$ Cholec $<$ Robust-MIS) as the definition of instrument (and/or counterexample) changes along with other factors.

\subsection{Implementation}
\label{sec:implementation} 
We have redesigned the implementation of the \emph{Endo-Sim2Real} framework in view of a teacher-student approach. To ensure a direct and fair comparison, we employ the same \textit{TerNaus11}~\cite{shvets2018automatic} as a backbone segmentation model.
Also, we utilize the best performing perturbation scheme (i.e. applying one of the \textit{pixel-intensity} perturbation\footnote{\textit{pixel-intensity}: random brightness and contrast shift, posterisation, solarisation, random gamma shift, random HSV color space shift, histogram equalization and contrast limited adaptive histogram equalization}
followed by one of the \textit{pixel-corruption} perturbation\footnote{\textit{pixel-corruption}: gaussian noise, motion blurring, image compression, dropout, random fog simulation and image embossing}) and the loss function (i.e. \emph{cross-entropy} and \emph{jaccard}) of \emph{Endo-Sim2Real} for evaluation.
All simulated input images and labels are first pre-processed with a stochastically-varying circular outer mask to give them the appearance of real endoscopic images.

We use a batch size of 8 for 50 epochs and apply weight decay ($1e-6$) as standard regularization.
During consistency training, we use a time-dependent weighting function, where the weight of the unlabeled loss term is linearly increased over the training.
The teacher model is updated with $\alpha$ (0.95) at each training step.

During evaluation of a dataset, we use an image-based dice score and average over all images to obtain a global dice metric for the dataset. 
For computation of the dice score, we exclude the cases where both the prediction and ground truth images are empty. However, we include cases with false positives for the empty images and set it to zero. So the dice score for empty ground-truth images (without any instrument) is either zero and considered in case of any false positives or undefined and not considered in case of correct prediction.
Also, we report all the results as an average performance of three runs throughout our experiments.

%% file: Tables/Datasets.tex
	\centering
	\setlength{\tabcolsep}{9pt} 
	\captionsetup{justification=centering}
	\caption{List of source (simulation) and target (real) datasets used during evaluation, where [videos (\#) $|$ empty frames (\%)] reflects the number of videos and percentage of frames with no instrument respectively}
	\label{table:datasets}
	\begin{tabular}{ c  c  c  c }
	    \hline\noalign{\smallskip}
    	\textbf{Dataset} & \textbf{Training} & \textbf{Testing} & \textbf{Instruments} \\
    	\noalign{\smallskip}\hline\noalign{\smallskip}
    	Simulation  & 20,000 (10 $|$ 33 \%)    &   n.a.       & two \\
    	\noalign{\smallskip}\hline\noalign{\smallskip}
    	Cholec15        & 5,034 (10 $|$ 12 \%)     &   2,136 (5 $|$ 13 \%)  & six {\footnotesize +} specimenbag \\
    	EndoVis         & 160 (4 $|$ 0 \%)        &   140 (6 $|$ 0 \%)    & seven \\
    	\noalign{\smallskip}\hline\noalign{\smallskip}
    	RobustMIS      & 5,983 (16 $|$ 17 \%)      &   4,057 (14 $|$ 20 \%)   & six $+$ trocar \\
    	Stage 1        &                           &   663 (2 $|$ 15 \%)      & - \\
    	Stage 2        &                           &   514 (2 $|$ 13 \%)      & - \\
    	Stage 3        &                           &   2880 (10 $|$ 23 \%)    & - \\
	    \noalign{\smallskip}\hline
    \end{tabular}

%% file: Sections/Results.tex

This section provides a quantitative comparison with respect to the state-of-the-art approaches to demonstrate the effectiveness of our approach.
Moreover, we perform a quantitative and qualitative analysis on three different datasets with varying degrees of the domain gap. This shows the strengths and weaknesses of our approach in order to better understand the challenges and provide valuable insights into addressing the remaining performance gap.

\begingroup
\begin{table}[b] 
	\subimport{./Tables/}{Quantitative_Comparison.tex}%
\end{table}
\endgroup

\subsection{Comparison with \emph{baseline} and \emph{state-of-the-art}}

In these experiments, we first highlight the performance of the two baselines: the \textit{lower baseline} (supervised learning  purely on simulated data) and the \textit{upper baseline} (supervised learning purely on annotated real data) in Table~\ref{table:Result_Compare}. 
The substantial performance gap between the baselines indicates the domain gap between simulated and real data.
Secondly, we compare our proposed teacher-student approach with other unsupervised domain adaptation approaches, i.e. the domain style transfer approach (\emph{I2I}) and the plain consistency-based joint learning approach (\emph{Endo-Sim2Real}) on the \textit{Cholec} dataset. The empirical results show that \emph{Endo-Sim2Real} works similar to \emph{I2I}, while our proposed approach outperforms both of these approaches.
Later, we evaluate our approach on two additional datasets and show that it consistently outperforms \emph{Endo-Sim2Real}.
These experiments demonstrate that the generalization performance of the DNN can be enhanced by employing unsupervised consistency learning on unlabeled data.
Finally, the performance gap with the upper baseline calls for identification of the issues needed to bridge the remaining domain gap.

\begingroup
\begin{figure}[b]
    \centering
    \includegraphics[width=1.0\textwidth]{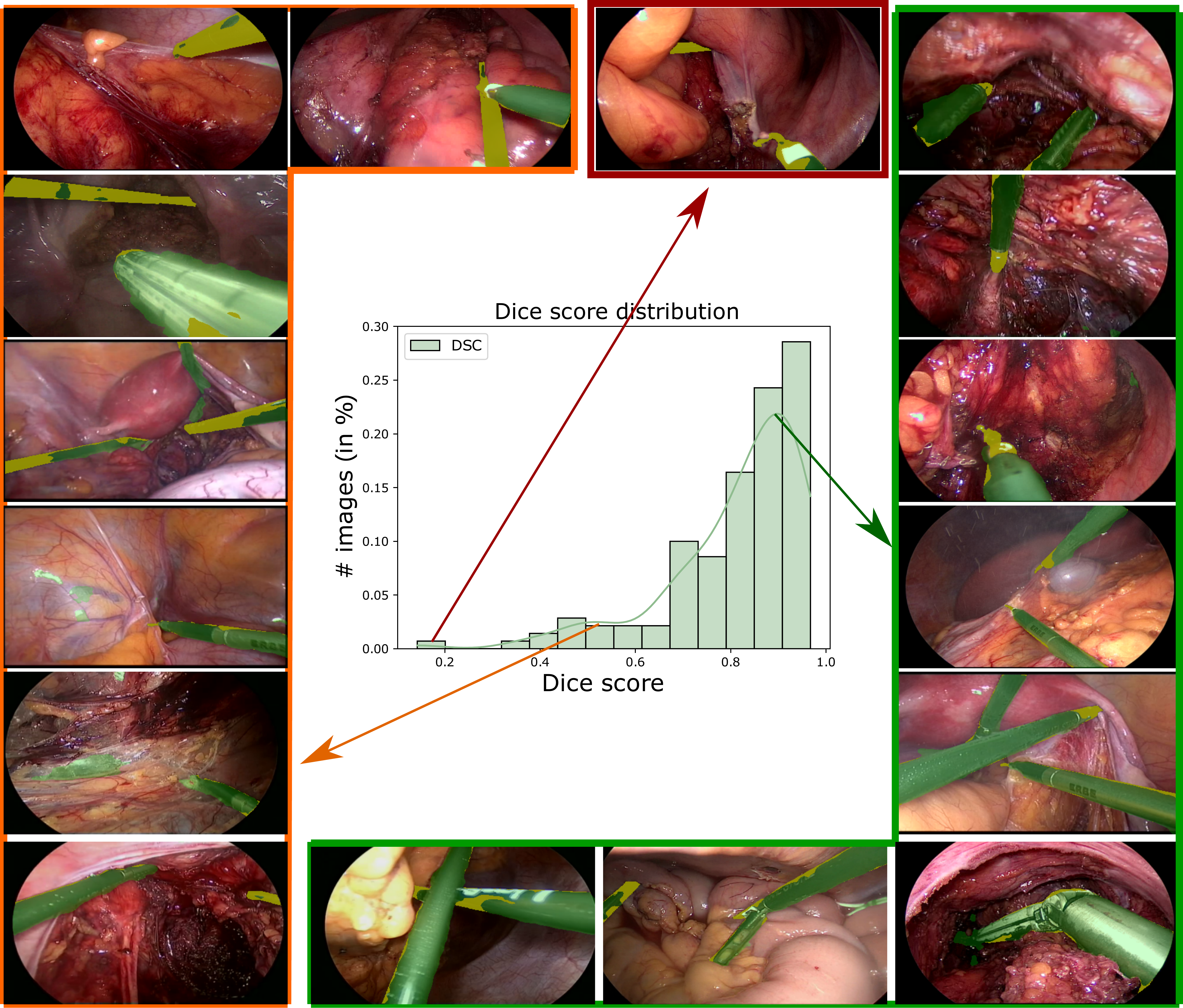}
	\caption{Qualitative analysis on EndoVis dataset. The green color in the images represents the network predictions while the yellow color represents under-segmentation.}
    \label{fig:qualitative_analysis_endovis}
\end{figure}
\endgroup

\subsection{Analysis on EndoVis}

Among the three datasets, our proposed approach performs best for EndoVis as shown in Table~\ref{table:Result_Compare}.
A visual analysis of the low performing cases in Figure~\ref{fig:qualitative_analysis_endovis} highlights factors such as false detection on specular reflection, under-segmentation for small instruments, tool-tissue interaction and partially occluded instruments.
These factors can in part be addressed by utilizing the temporal information of video frames.~\cite{gonzalez2020isinet}

\begingroup
\begin{landscape}
\begin{figure}[htbp]
    \centering
	\includegraphics[width=0.95\columnwidth]{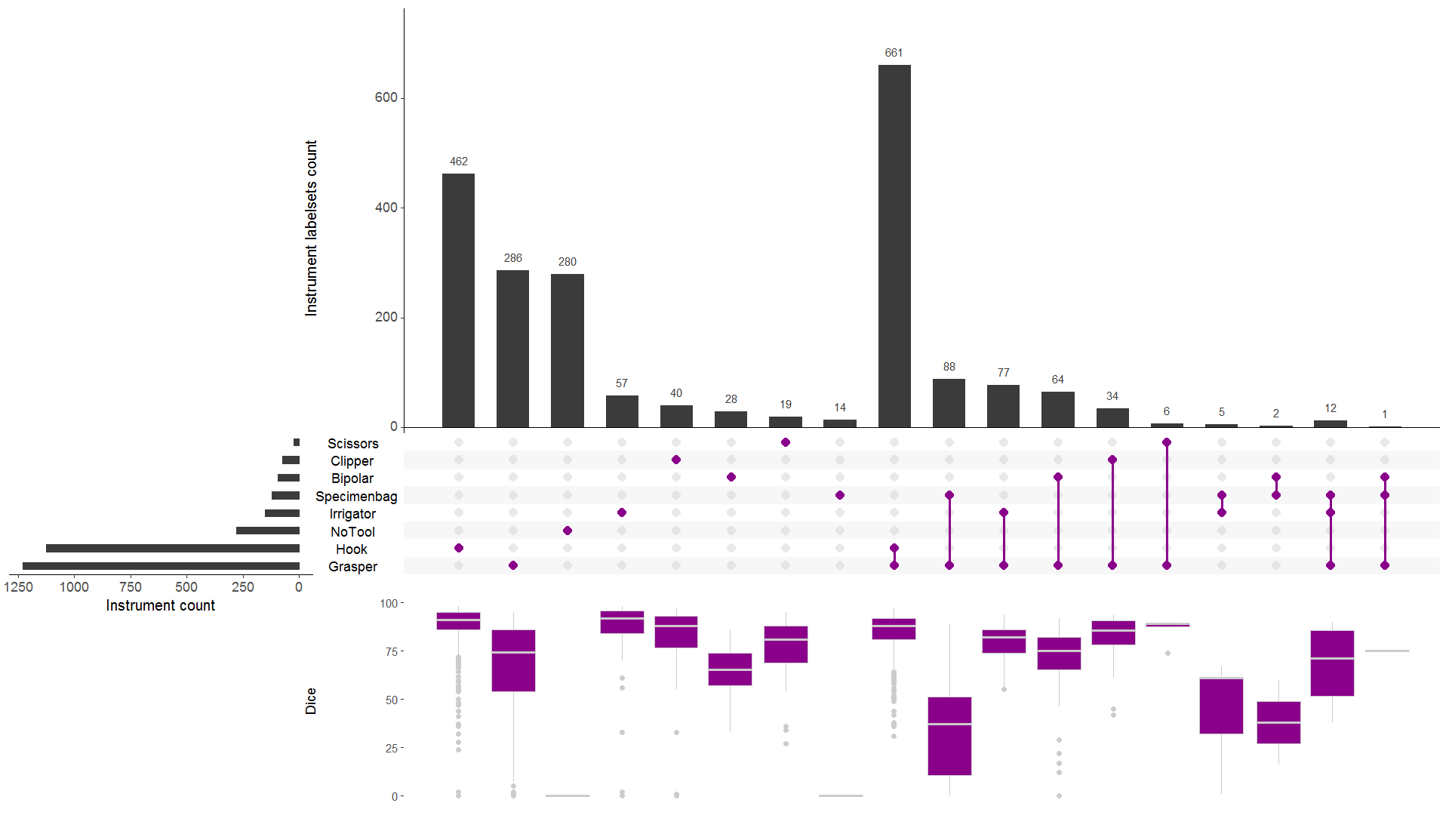}
	\caption{Visualization of the relation between tool co-occurrence and segmentation quality for the Cholec dataset. Please note that the dice score is zero for no tool cases and specimen bag as it is treated as background.}
    \label{fig:upset}
\end{figure}
\end{landscape}
\endgroup

\subsection{Analysis on Cholec}

We performed an extensive performance analysis of our proposed approach on the Cholec dataset as instrument-specific labels are available for it (in comparison to EndoVis).
To understand the distinctive performance aspects for the Cholec dataset, we compare the segmentation performance across different instrument co-occurrence in Figure~\ref{fig:upset}.
A similar range of dice scores highlights that the performance of our approach is less impacted by the presence of multiple tool combinations in an endoscopic image.
However, it also clearly shows that the segmentation performance of our approach drops when the specimen bag and its related co-occurrences are present (as seen in the respective box plots in Figure~\ref{fig:upset}). A visual analysis highlights false detection on the reflective surface of the specimen bag.

\begingroup
\begin{figure}[htbp]
    \centering
    \includegraphics[width=1.0\textwidth]{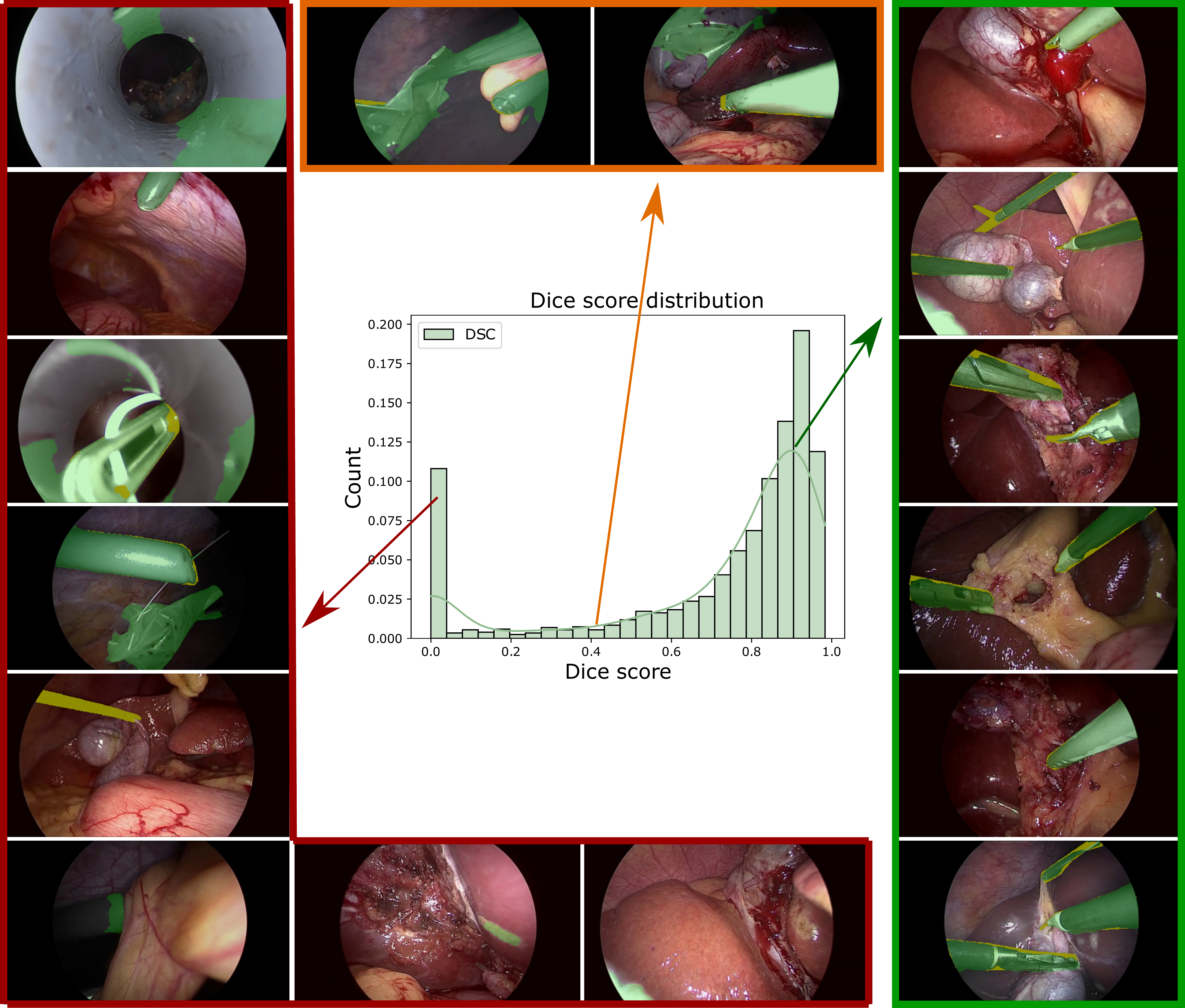}
	\caption{Qualitative analysis on the Cholec dataset. The green color in the images represents the network predictions while the yellow color represents under-segmentation.}
    \label{fig:qualitative_analysis_cholec}
\end{figure}
\endgroup

Apart from the previously analyzed performance degrading factors in the EndoVis dataset, other major factors affecting the performance are as follows:
\begin{itemize}
  \item[$\ast$] \textbf{Out of distribution cases} such as a non-conventional tool-shape-like instrument: specimen bag (see box-plots for labelsets with specimen bag in Figure~\ref{fig:upset}).
  \item[$\ast$] \textbf{False detection for scenarios} such as an endoscopic view within the trocar, instrument(s) near the image border or under-segmentation for small instruments.
  \item[$\ast$] \textbf{Artefact cases} such as specular reflection. The impact of other artefacts such as blood, smoke or motion blur is lower.
\end{itemize}

\noindent Although our proposed approach struggles to tackle these artefacts and out of distribution cases, addressing these performance degrading factors is itself an open research problem.~\cite{ali2020artefacts}

\begingroup
\begin{figure}[htbp]
    \centering
    \includegraphics[width=1.0\textwidth]{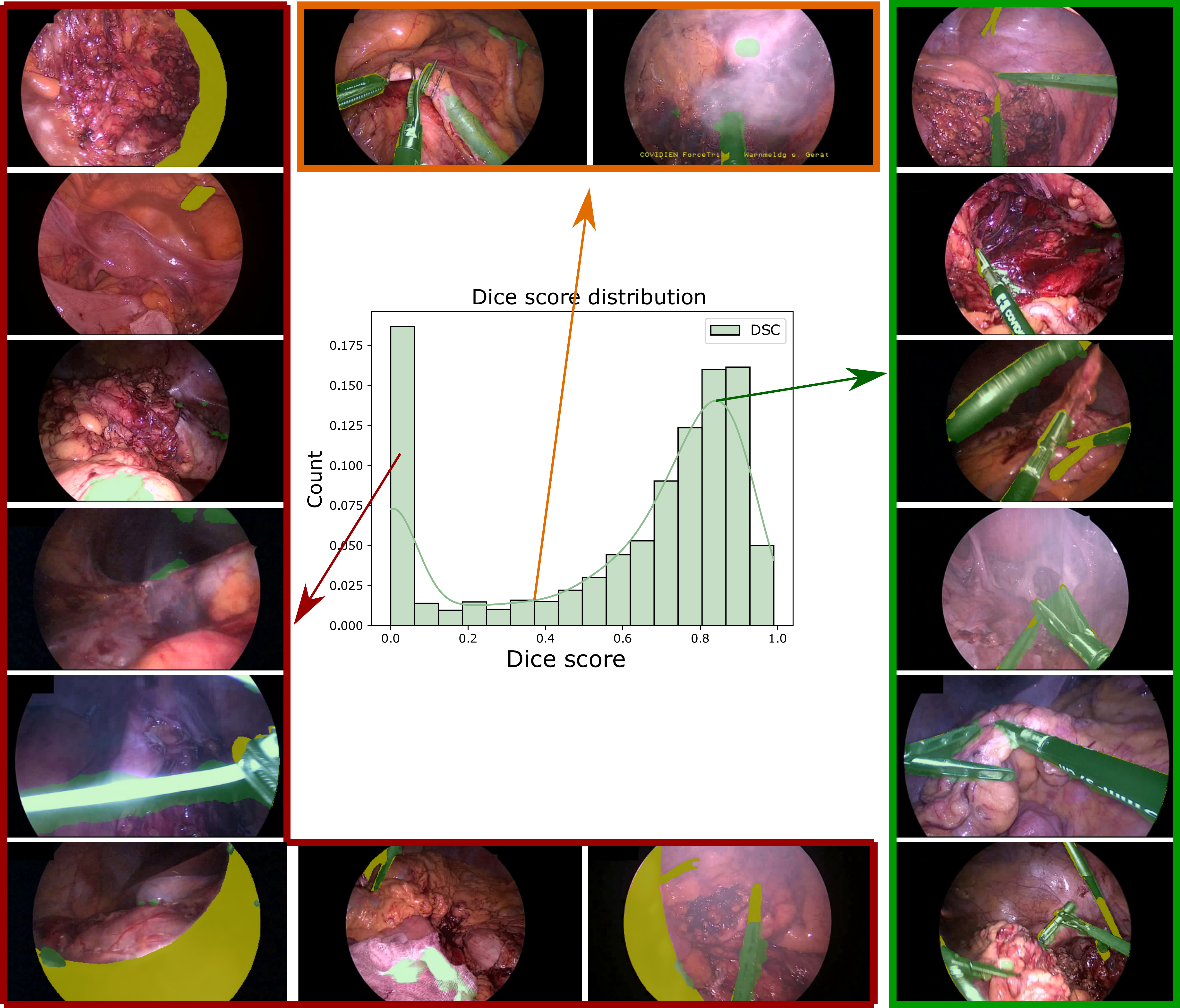}
	\caption{Qualitative analysis on the RobustMIS dataset. The green color in the images represents the network predictions while the yellow color represents under-segmentation.}
    \label{fig:qualitative_analysis_rs}
\end{figure}
\endgroup

\begingroup
\begin{table}[htbp] 
	\subimport{./Tables/}{Quants_RobustMIS.tex}%
\end{table}
\endgroup

\subsection{Analysis on RobusMIS}

We analyzed the performance of our approach on images with a different number of instruments in the RobustMIS dataset. We found that the performance is not significantly affected by the presence of multiple tools (see Table ~\ref{table:dice_per_tool}).
A low performance for a single visible instrument is attributed to small, stand-alone instruments across image boundary.
Apart from the factors in the Cholec dataset, other real-world performance degrading factors in RobustMIS include: presence of other out-of-distribution cases such as non-rigid tubes, bandages, needles etc.; presence of corner cases such as trocar-views and specular reflections producing a instrument-shape-like appearance.
These failure cases highlight a drawback of our approach, which works under the assumption that the shape of the instrument remains consistent between the domains. Therefore, our approach may not be able to produce faithful predictions in case instruments with different shapes are encountered in the real domain (compared to instruments in simulation) or counterexamples with instrument like appearance.

\subsection{Impact of empty ground-truth frames}

The performance of our teacher-student approach is negatively affected by the video frames that do not contain instruments.
This is because the dice score is assigned to zero when the network predicts false positives (as seen in Figure~\ref{fig:qualitative_analysis_cholec} and ~\ref{fig:qualitative_analysis_rs}) in instrument-free video frames.
A direct relation of this effect can be seen in Table~\ref{table:Result_Compare} where the dice score across the datasets decreases as the number of empty frames increases (in \%) from EndoVis to RobustMIS. 
It suggests that utilizing false detection techniques in the current framework can help in enhancing the generalization capabilities.

%% file: Tables/Quantitative_Comparison.tex
	\centering
	\captionsetup{justification=centering}
	\caption{Quantitative comparison using DSC [mean (std)]. The two baselines: simulation only and real only means training only on the simulated data and training on the annotated real data respectively (i.e. no adaptation).
	The paired t-test for \textit{Endo-Sim2Real} and our work results in \textit{p-value} $<<$ $0.01$.}
	\label{table:Result_Compare}
	\begin{tabular}{ c  c  c  c  c  c }
    	\hline\noalign{\smallskip}
    	Approach & EndoVis & Cholec &  R-MIS (S 1)  & R-MIS (S 2)  & R-MIS (S 3) \\
    	\noalign{\smallskip}\hline\noalign{\smallskip}
    	Simulation only      & .42 (.29)   & .30 (.30) &  .30 (.28) & .36 (.30) & .20 (.24) \\
    	\texttt{I2I}~\cite{pfeiffer2019generating}   & n.a. & .68 (.30) &  n.a. & n.a. & n.a. \\
    	Endo-Sim2Real~\cite{sahu2020endo}    & .76 (.17) & .68 (.31) &  .57 (.33) & .60 (.32) & .51 (.35) \\
    	Teacher-Student      & \textbf{.80} (.16) & \textbf{.72} (.30) &  \textbf{.61} (.32) & \textbf{.65} (.31) & \textbf{.58} (.34) \\
    	Real only           & .93 (.06) & .86 (.24) &  .82 (.27) & .83 (.25) & .78 (.30) \\
    	\noalign{\smallskip}\hline\noalign{\smallskip}
    	\textcolor{gray}{Empty frames}           & \textcolor{gray}{0 \%} & \textcolor{gray}{13 \%} &  \textcolor{gray}{15 \%} & \textcolor{gray}{13 \%} & \textcolor{gray}{23 \%} \\
	    \noalign{\smallskip}\hline
    \end{tabular}

%% file: Tables/Quants_RobustMIS.tex
	\centering
	\captionsetup{justification=centering}
	\caption{Quantitative results for multi-instrument presence in RobustMIS dataset using DSC [mean (std)]}
	\label{table:dice_per_tool}
	\begin{tabular}{ c  c  c  c  c  c  c  c  c }
    	\hline\noalign{\smallskip}
    	No. & 1 & 2  & 3 & 4 & 5 & 6 & 7 \\
    	\noalign{\smallskip}\hline\noalign{\smallskip}
    	\#   & 1802  & 1173  & 229   & 31    & 3     & -     & 1 \\
    	DSC  &  .61 (.34)  & .72 (.22)   & .73 (.18)   & .76 (.11)   & .76 (.13)   & -     & .77 (.00) \\
	    \noalign{\smallskip}\hline
    \end{tabular}

%% file: Sections/Discussion.tex
We introduce teacher-student learning to address the confirmation bias issue of the \emph{EndoSim2Real} consistency learning.
This enables us to tackle the challenging problem of the domain shift between synthetic and real images for surgical tool segmentation in endoscopic videos.
Our proposed approach enforces the teacher model to generate reliable targets to facilitate stable student learning.
Since the teacher is a moving average model of the student, the extension does not add computational complexity to the current approach.

We show that the proposed teacher-student learning approach generalizes across three different datasets for the instrument segmentation task and consistently outperforms the previous state-of-the-art.
For a majority of images (see high peak in Figure~\ref{fig:qualitative_analysis_endovis}, ~\ref{fig:qualitative_analysis_cholec} and ~\ref{fig:qualitative_analysis_rs}), the segmentation predictions are usually correct with small variations across the instrument boundary.
Moreover, a thorough analysis of the results highlight interpretable failure modes of simulation-to-real deep learning as the domain gap widens progressively.

Considering the strengths and weaknesses of our teacher-student enabled consistency-based unsupervised domain adaptation approach, the framework admits multiple straightforward extensions:
\begin{itemize}
  \item[$\ast$] Implementing techniques to suppress false detection for empty frames, instruments near the image border and specular reflections, for instance by utilizing temporal information~\cite{gonzalez2020isinet} of video frames.
  \item[$\ast$] Improving physical properties of simulation to capture instrument-tissue interaction, considering the variations in predictions across instrument boundaries.
  \item[$\ast$] Extension towards other domain-adaptation-tasks, for instance depth estimation\cite{mahmood2018unsupervised,rau2019implicit} or instrument pose estimation\cite{du2018articulated2dpose,allan2018articulated3dpose} by exploiting depth maps from the simulated virtual environments.
  \item[$\ast$] Extension towards semi-supervised domain adaptation or real-to-real unsupervised domain adaptation by joint learning from labeled and unlabeled data.
  \item[$\ast$] Utilizing this approach on top of self-supervised or adversarial domain mapping approaches such as $I2I$.
\end{itemize}

\noindent The heavy reliance of current approaches on manual annotation and the harsh reality of surgeons sparing time for the annotation process propels simulation-to-real domain adaptation as the obvious problem to address in surgical data science.
The proposed approach ushers annotation-efficient surgical data science for the operating room of the future.

%% file: Sections/COI.tex
\section*{Author Statement}
\textbf{Conflict of interest}: The authors state no conflict of interest.
\smallskip
\\ \textbf{Informed consent}: This study contains patient data from a publicly available dataset.
\smallskip
\\ \textbf{Ethical approval}: This article does not contain any studies with human participants or animals performed by any of the authors.

